\documentclass{article}
\usepackage[utf8]{inputenc}
\usepackage[a4paper]{geometry}

\usepackage{times}
\usepackage{epsfig}
\usepackage{microtype}
\usepackage{graphicx}
\usepackage{amsmath}
\usepackage{amssymb}
\usepackage{adjustbox}
\usepackage{algorithm}
\usepackage{algorithmic}
\usepackage{url}
\usepackage{float}

\title{Learning From Less Data: Diversified Subset Selection and Active Learning in Image Classification Tasks}

\author{Vishal Kaushal\\
IIT Bombay\\
Mumbai, Maharashtra, India\\
{\tt\small vkaushal@cse.iitb.ac.in}
\and
Anurag Sahoo\\
AITOE Labs\\
Seattle, Washington, USA\\
{\tt\small anurag@aitoelabs.com}
\and
Khoshrav Doctor\\
AITOE Labs\\
Mumbai, Maharashtra, India\\
{\tt\small khoshrav@gmail.com}
\and
Narsimha Raju\\
IIT Bombay\\
Mumbai, Maharashtra, India\\
{\tt\small uavnraju@cse.iitb.ac.in}
\and
Suyash Shetty\\
AITOE Labs\\
Mumbai, Maharashtra, India\\
{\tt\small suyashshetty29@gmail.com}
\and
Pankaj Singh\\
IIT Bombay\\
Mumbai, Maharashtra, India\\
{\tt\small pr.pankajsingh@gmail.com}
\and
Rishabh Iyer\\
AITOE Labs\\
Seattle, Washington, USA\\
{\tt\small rishabh@aitoelabs.com}
\and
Ganesh Ramakrishnan\\
IIT Bombay\\
Mumbai, Maharashtra, India\\
{\tt\small ganesh@cse.iitb.ac.in}
}

\begin{document}

\maketitle

\begin{abstract}
Supervised machine learning based state-of-the-art computer vision techniques are in general data hungry and pose the challenges of not having adequate computing resources and of high costs involved in human labeling efforts. Training data subset selection and active learning techniques have been proposed as possible solutions to these challenges respectively. A special class of subset selection functions naturally model notions of diversity, coverage and representation and they can be used to eliminate redundancy and thus lend themselves well for training data subset selection. They can also help improve the efficiency of active learning in further reducing human labeling efforts by selecting a subset of the examples obtained using the conventional uncertainty sampling based techniques. In this work we empirically demonstrate the effectiveness of two diversity models, namely the Facility-Location and Disparity-Min models for training-data subset selection and reducing labeling effort. We do this for a variety of computer vision tasks including Gender Recognition, Scene Recognition and Object Recognition. Our results show that subset selection done in the right way can add 2-3\% in accuracy on existing baselines, particularly in the case of less training data. This allows the training of complex machine learning models (like Convolutional Neural Networks) with much less training data while incurring minimal performance loss.
\end{abstract}

\section{Introduction}
Deep Convolutional Neural Network based models are today the state-of-the-art for most computer vision tasks. Seeds of the idea of deep learning were sown around the late 90's ~\cite{lecun1998gradient} and it gained popularity in 2012 when AlexNet ~\cite{krizhevsky2012imagenet} won the challenging ILSVRC (ImageNet Large-Scale Visual Recognition Challenge) 2012 competition ~\cite{russakovsky2015imagenet}, 
demonstrating an astounding improvement over the then state-of-the-art image classification techniques. AlexNet reported a top-5 test error rate of 15.4\%. This was soon followed by an upsurge of deep models for computer vision tasks from all over the community. ZFNet ~\cite{zeiler2014visualizing}, the winner of ILSVRC 2013 improved upon AlexNet to bring down the top-5 test error rate to 11.2\%. Through their novel visualization techniques, they provided good intuition to the workings on CNNs and illustrated more ways to improve performance. They argued that this renewed interest (since they were proposed in 1998 ~\cite{lecun1998gradient}) in CNNs is due to the accessibility of large training sets and increased computational power, thanks to GPUs. VGGNet ~\cite{simonyan2014very} demonstrated that a simpler, but deeper model can be used to bring the error further down to 7.3\%. Their work reinforced the idea that Convolutional Neural Networks have to be deep enough for the hierarchical representation of visual data to be effective. Then came the even deeper GoogLeNet ~\cite{szegedy2015going}, the winner of ILSVRC 2014, that became the first CNN to be fundamentally architecturally different from AlexNet. GoogLeNet introduced the idea that CNN layers didn't always have to be stacked up sequentially. ResNet ~\cite{he2016deep} was even deeper, a phenomenal 152 layer deep network and won ILSVRC 2015 with an incredible error rate of 3.57\%, beating humans in the image classification task. 

Similarly, for object detection tasks, the first significant advancement in deep learning was made by RCNNs ~\cite{girshick2014rich}, which was followed by Fast RCNN ~\cite{girshick2015fast} and then Faster RCNN ~\cite{ren2015faster} for performance improvement. More recently, YOLO ~\cite{redmon2016you} and YOLO9000 ~\cite{redmon2016yolo9000} have emerged as the state-of-the-art in object detection. YOLO's architecture is inspired by GoogLeNet and consists of 24 convolutional layers followed by two fully connected layers. State-of-the-art face recognition techniques such as DeepFace ~\cite{taigman2014deepface}, DeepID3 ~\cite{sun2015deepid3}, Deep Face Recognition ~\cite{parkhi2015deep} and FaceNet ~\cite{schroff2015facenet} also consist of deep convolutional networks. Similar is the story with state-of-the-art scene recognition techniques ~\cite{zhou2014learning}, and other computer vision tasks such as age and gender classification~\cite{levi2015age} etc.

Every coin, however, has two sides and so is the case with this golden coin of deep learning. While deeper models are increasingly improving for computer vision tasks, they pose challenges of increased training complexity, enormous data requirements and high labeling costs. 

Training complexity and huge data requirements is due to the depth of the network and large number of parameters to be learnt. A large deep neural net with a large  number of parameters to learn (and hence a large degree of freedom) has a very complex and extremely non-convex error surface to traverse and thus it requires a great deal of data to successfully search for a reasonably optimal point on that surface. AlexNet, for example was trained on ImageNet data ~\cite{deng2009imagenet}, which contained over 15 million annotated images from a total of over 22,000 categories. It was trained on two GTX 580 GPUs for five to six days. ZFNet used \emph{only} 1.3 million images and trained on a GTX 580 GPU for twelve days. VGGNet has 140 million parameters to learn and was trained on 4 Nvidia Titan Black GPUs for two to three weeks. GoogLeNet was trained on ``a few high-end GPUs within a week''. ResNet was trained on an 8 GPU machine for two to three weeks. 

One way researchers have addressed this challenge is through architectural modifications to the network. For example, by making significant architectural changes, GoogLeNet improved utilization of the computing resources inside the network thus allowing for increased depth and width of the network while keeping the computational budget constant. Similarly, by explicitly reformulating the layers as learnt residual functions with reference to the layer inputs, instead of learning unreferenced functions, ResNet allows for easy training of much deeper networks. Highway networks ~\cite{srivastava2015training} introduce a new architecture allowing a network to be trained directly through simple gradient descent. With the goal of reducing training complexity, some other studies ~\cite{ba2014deep, hinton2015distilling, iandola2016squeezenet} have also focused on model compression techniques. On similar lines, studies like ~\cite{levi2015age} propose simpler network architectures in favor of availability of limited training data. In this work, however, we are \emph{not}  addressing this approach. 

Other approaches advocate use of pre-trained models, trained on large data sets and presented as `ready-to-use' but they rarely work out of the box for domain specific computer vision tasks (such as detecting security sensitive objects). It thus becomes important to create customized models, which again potentially require significant amount of training data. Domain adaptation techniques like transfer learning and fine tuning allow creation of customized models with much lesser data. Transfer learning allows models trained on one task to be easily reused for a different task ~\cite{pan2010survey}. Several studies ~\cite{donahue2014decaf, yosinski2014transferable, sharif2014cnn, oquab2014learning,long2015learning} have analyzed the transferability of features in deep convolutional networks across different computer vision tasks. It has been demonstrated that transferring features even from distant tasks can be better than using random features and that deep convolutional features can be the preferred choice in most visual recognition tasks. In a yet another interesting approach for reducing data requirements, Spatial Transfer Networks ~\cite{jaderberg2015spatial} explore the idea of transforming the image (and not the architecture) to make the model invariant to images with different scales and rotations which could have otherwise been achieved only by a lot more data. In our work, we demonstrate the utility of subset selection methods in alleviating this big data problem.

Orthogonal to the challenge of training complexity is the challenge of unavailability of labeled data. Human labeling efforts are costly and grow exponentially with the size of the dataset ~\cite{vondrick2013efficiently}. Again, several approaches have been proposed to address this problem. In ~\cite{oquab2015object} the authors describe a weakly supervised convolutional neural network (CNN) for object classification that relies only on image-level labels, yet can learn from cluttered scenes containing multiple objects to produce their approximate locations. Zero-shot learning ~\cite{socher2013zero, changpinyo2016synthesized} and one-shot learning ~\cite{vinyals2016matching} techniques also help address the issue of lack of annotated examples. On the other hand, active learning approaches try to minimize the labeling costs. The core idea behind active learning ~\cite{settles2010active} is that a machine learning algorithm can achieve greater accuracy with fewer training labels if it is allowed to choose the data from which it learns. In our work we demonstrate the utility of various subset selection functions in active learning setting thereby identifying an even smaller set of data points to label for computer vision tasks, yet incurring minimal performance loss.

\subsection{Existing Work}
A common approach for training data subset selection is to use the concept of a coreset ~\cite{agarwal2005geometric}, which aims to efficiently approximate various geometric extent measures over a large set of data instances via a small subset. Submodular functions naturally model the notion of diversity, representation and coverage and hence submodular function optimization has been applied to recommendation systems to recommend relevant and diverse items which explicitly takes the coverage of user interest into account ~\cite{shaframework}. It has also been applied to data summarization and data partitioning ~\cite{wei2016submodular}.Training data subset selection has been studied as a problem of submodular maximization, where appropriate submodular functions are chosen to measure the utility of each subset for an underlying task ~\cite{shinohara2014submodular, zheng2014submodular, prasad2014submodular}.

There have been very few studies in active learning and subset selection for Computer Vision tasks. In ~\cite{gavves2015active}, Gavves et al combine techniques from transfer learning, active learning and zero-shot learning to reuse existing knowledge and reduce labeling efforts. Zero-shot learning techniques don't expect any annotated samples, unlike active learning and transfer learning which assume either the availability of at least a few labeled target training samples or an overlap with existing labels from other datasets. By using a max-margin formulation (which transfers knowledge from zero-shot learning models used as priors) they present two basic conditions to choose greedily only the most relevant samples to be annotated, and this forms the basis of their active learning algorithm. Using information theoretic objective function such as mutual information between the labels, Sourati et al have developed a framework ~\cite{sourati2016classification} for evaluating and maximizing this objective for batch mode active learning. Another recent study has adapted batch active learning to deep models ~\cite{ducoffescalable}. Most batch-active learning techniques involve the computation of Fisher matrix which is intractable for deep networks. Their method relies on computationally tractable approximation of the Fisher matrix, thus making them relevant in the context of deep learning.

Our work is more closely related to the work of ~\cite{wei2015submodularity} which has shown the mathematical connection of submodularity to the data likelihood functions for Naive Bayes and Nearest Neighbor classifiers, thereby allowing for formulation of the data subset selection problems  for  these  classifiers  as  constrained submodular  maximization. They extend this framework to active learning, combining  the  uncertainty  sampling  method  with  a submodular data subset selection framework. They argue that while traditional active learning techniques aim at labeling the most informative instances, these methods neglect the notion of representativeness which can be useful to reduce the size of the subset further by eliminating redundancy. Their work is mainly theoretical and was applied in the context of text recognition. This work attempts to study the role of subset selection in computer vision tasks, where we apply the framework of submodular subset selection and active learning to large scale image recognition tasks. Our submodular subset selection framework naturally combines the notion of informativeness as well as representativeness. Our work can easily be extended to other query strategies like \emph{query by committee}. 

\subsection{Our Contributions}
Extending the work of ~\cite{wei2015submodularity}, in this work, we make three significant contributions:

\begin{enumerate}
\item We demonstrate the utility of subset selection in training deep models with a much smaller subset of training data, yet without significant compromise in accuracy for a variety of computer vision tasks. This is also particularly relevant for training a nearest neighbor model given that the complexity of inference of this non-parametric classifier is directly proportional to the number of training examples
\item We demonstrate the utility of subset selection in reducing labeling effort by eliminating the redundant instances amongst those chosen to be labeled by the traditional uncertainty method in every iteration of mini-batch adaptive active learning. 
\item We provide insights to the choice of two different diversity models for subset selection, namely Facility-Location and Disparity-Min. Both functions have very different characteristics, and we try to argue scenarios where we might want to use one over the other.
\item We empirically demonstrate the utility of this framework for several complex computer vision tasks including Gender Recognition, Object Recognition and Scene Recognition. Moreover, we apply this to Fine tuning Deep Models (end to end experiments) as well as Transfer Learning. 
\end{enumerate}

\section{Background}
\subsection{Techniques for diverse subset selection}
Given a set $V = \{1, 2, 3, \cdots, n\}$ of items which we also call the \emph{Ground Set}, define a utility function (set function) $f:2^V \rightarrow \mathbf{R}$, which measures how good a subset $X \subseteq V$ is. Let $c :2^V \rightarrow \mathbf{R}$ be a cost function, which describes the cost of the set (for example, the size of the subset). The goal is then to have a subset $X$ which maximizes $f$ while simultaneously minimizes the cost function $c$. It is easy to see that maximizing a generic set function becomes computationally infeasible as $V$ grows. Often the cost $c$ is budget constrained (for example, a fixed set summary) and a natural formulation of this is the following problem:
\begin{align}
\mbox{Problem 1: } \max\{f(X) \mbox{ such that } c(X) \leq b\}
\end{align}

A special class of set functions, called submodular functions ~\cite{nemhauser1978analysis}, however, makes this optimization easy. Submodular functions exhibit a property that intuitively formalizes the idea of ``diminishing returns''. That is, adding some instance $x$ to the set $A$ provides more gain in terms of the target function than adding $x$ to a larger set $A'$, where $A \subseteq A'$.  Informally, since $A'$ is a superset of $A$ and already contains more information, adding $x$ will not help as much. Using a greedy algorithm to optimize a submodular function (for selecting a subset) gives a lower-bound performance guarantee of around 63\% of optimal ~\cite{nemhauser1978analysis} to Problem 1, and in practice these greedy solutions are often within 90\% of optimal ~\cite{krause2008optimizing}. This makes it advantageous to formulate (or approximate) the objective function for data selection as a submodular function. 

Several diversity and coverage functions are submodular, since they satisfy this diminishing returns property. The ground-set $V$ and the items $\{1, 2, \cdots, n\}$ depend on the choice of the task at hand. Submodular functions have been used for several summarization tasks including Image summarization~\cite{tschiatschek2014learning}, video summarization~\cite{gygli2015video}, document summarization~\cite{lin2011class}, training data summarization and active learning~\cite{wei2015submodularity} etc.

Building on these lines, in this work we demonstrate the utility of subset selection in allowing us to train machine learning models using a subset of training data without significant loss in accuracy. In particular, we focus on a) \emph{Facility-Location} function, which models the notion of representativeness and b) \emph{Disparity-Min} function, which models the notion of diversity.

Representation based functions attempt to directly model representation, in that they try to find a representative subset of items, akin to centroids and medoids in clustering. The Facility-Location function ~\cite{mirchandani1990discrete} is closely related to k-medoid clustering. Denote $s_{ij}$ as the similarity between images $i$ and $j$. We can then define $f(X) = \sum_{i \in V} \max_{j \in X} s_{ij}$. For each image $i$, we compute the representative from $X$ which is closest to $i$ and add the similarities for all images. Note that this function, requires computing a $O(n^2)$ similarity function. However, as shown in~\cite{wei2014fast}, we can approximate this with a nearest neighbor graph, which will require much less storage, and also can run much faster for large ground set sizes.

Diversity based functions attempt to obtain a diverse set of keypoints. The goal is to have minimum similarity across elements in the chosen subset by maximizing minimum pairwise distance between elements. There is a subtle difference between the notion of diversity and the notion of representativeness. While diversity \emph{only} looks at the elements in the chosen subset, representativeness also worries about their similarity with the remaining elements in the superset. Denote $d_{ij}$ as a distance measure between images $i$ and $j$. Define a set function $f(X) = \min_{i, j \in X} d_{ij}$. This function, called Disparity-Min is not submodular, but can be efficiently optimized via a greedy algorithm~\cite{dasgupta2013summarization}. It is easy to see that maximizing this function involves obtaining a subset with maximal minimum pairwise distance, thereby ensuring a diverse subset of snippets or keyframes. 

Another example of diversity based functions are Determinantal Point Processes, defined as $p(X) = \mbox{Det}(S_X)$ where $S$ is a similarity kernel matrix, and $S_X$ denotes the rows and columns instantiated with elements in $X$. It turns out that $f(X) = \log p(X)$ is submodular, and hence can be efficiently optimized via the Greedy algorithm. However, unlike the above functions, this requires computing the determinant and is $O(n^3)$ where $n$ is the size of the ground set. This function is not computationally feasible and hence we do not consider it in our experiments.

\subsection{Techniques for active learning}
Active learning can be implemented in three flavors ~\cite{guillory2012active}:
\begin{enumerate}
\item Batch active learning - there is one round of data selection and the data points are chosen to be labeled without  any  knowledge  of  the  resulting  labels  that  will be returned
\item Adaptive active learning - there are many rounds of data selection, each of which selects one data point whose label may be used to select the data point at future rounds
\item Mini-batch adaptive active learning - a hybrid scheme where in each round a mini-batch of data points are selected to be labeled, and that may inform the determination of future mini-batches. In our work, this flavor of active learning is what we focus on.
\end{enumerate}

In this work, we focus on the third flavor, i.e. mini-batch adaptive active learning. As far as the query framework in active learning is concerned, the simplest and most commonly used query framework is uncertainty sampling ~\cite{lewis1994sequential} wherein an active learner queries the instances about which it is least certain how to label. Getting a label of such an instance from the oracle would be the most informative for the model. Another,  more theoretically-motivated query selection framework is the query-by-committee (QBC) algorithm ~\cite{seung1992query}. In this approach, a committee of models is maintained which are all trained on the current labeled set but represent competing hypothesis. Each committee member is then allowed to vote on the labelings of query candidates.  The most informative query is considered to be the instance about which they most disagree. There are other frameworks like Expected Model Change ~\cite{settles2008multiple}, Expected Error Reduction ~\cite{roy2001toward}, Variance Reduction ~\cite{cohn1994neural} and Density-Weighted methods ~\cite{settles2008analysis}. For a more detailed analysis of active learning in general, readers are encouraged to read a survey by Settles ~\cite{settles2010active}. Building upon the ideas developed by Wei et al ~\cite{wei2015submodularity}, in this work we demonstrate the utility of submodular functions in identifying an even smaller set of data points to label for computer vision tasks, yet incurring minimal performance loss.
\section{Methodology}
In line with our goals of empirically establishing the utility of subset selection in two different settings, we devise two different sets of experiments (Goal-1 \& Goal-2) on various data sets for a variety of computer vision tasks. 

In Goal-1, we evaluate the efficiency of our subset selection methods to enable learning from lesser data, yet incurring minimal loss in performance. Towards this, we apply Facility-Location and Disparity-Min based subset selection techniques for training a nearest neighbor classifier (kNN). For a given dataset we run several experiments to evaluate the accuracy of kNN, each time trained on a different sized subset of training data. We take subsets of sizes from 5\% to 100\% of full training data, with a step size of 5\%. For each experiment we report the accuracy of kNN over the hold-out data. We also compare these results against kNN trained on randomly selected subsets. Concretely, denote $f$ as the measure of \emph{information} as to how well a subset $X$ of training data performs as a proxy to the entire dataset. We then model this as an instance of Problem 1 with $f$ being the Facility-Location and Disparity-Min Functions. 

In Goal-2, we evaluate the efficiency of our subset selection methods in the context of mini-batch adaptive active learning to demonstrate savings on labeling effort. We propose a Submodular Active Learning algorithm similar to~\cite{wei2015submodularity}. The idea is to use uncertainty sampling as a filtering step to remove less informative training instances. Denote $\beta$ as the size of the filtered set $F$. We then select a subset $X$, as
\begin{align}
\mbox{Problem 2: } \max\{f(X) \mbox{ such that } |X| \leq B, X \subseteq F\}
\end{align}
where $B$ is the number of instances selected by the batch active learning algorithm at every iteration. In our experiments, we use $f$ as Facility-Location and Disparity-Min functions. For a rigorous analysis and a fair evaluation, we implement this both in the context of transfer learning (which can be seen as a smart cold start, Goal-2b) - as well as fine tuning (which can be seen as a warm-start, Goal-2a). For transfer learning (Goal-2b), we extract the features from a pre-trained CNN relevant to the computer vision task at hand and train a Logistic Regression classifier using those features. 

The basic algorithm for Goal-2 experiments is as mentioned in the Algorithm \ref{Algo:Goal-2} 
\begin{algorithm}[H]
\caption{Goal-2: Submodular Active Learning}\label{Algo:Goal-2}
\begin{algorithmic}[1]
\STATE Start with a small initial seed labeled data, say $L$
\FOR{each round 1 to $T$ }
\STATE Fine tune a pre-trained CNN (Goal-2a) or train a Logistic Regression classifier (Goal-2b) with the labeled set $L$
\STATE Report the accuracy of this model over the hold-out data.
\STATE Using this model, compute uncertainties of remaining unlabeled data points $U$, and select a subset $F, F \subseteq U$ of the most uncertain data points. 
\STATE Solve Problem 2 with Facility-Location and Disparity-Min Functions 
\STATE Label the selected subset and add to labeled set $L$
\ENDFOR
\end{algorithmic}
\end{algorithm}

The uncertainties of unlabeled data points are computed in one of the following ways:
\begin{align*}
          u &= 1-\max_i p_i && \\
          u &= 1-(\max_i p_i- \max_{j \in \{C\}-\arg\max_i  p_i} p_j )\\
          u &= -\sum_i {p_i * \log_2 ( p_i )}\\
 \text{where:}\quad 
u &= \text{uncertainty}\\
p_i &= \text{probability of class $i$}\\
\{C\} &= \text{set of classes}\\
\end{align*}

The parameters $B$ and $\beta$ of FASS: $B$ represents the percentge of images that are to be labeled at the end of each round and added to the training set. $\beta$ specifies what percentage of data (sorted in decreasing order of their hypothesized uncertainties by the current model) forms the ground set (for subset selection) in every round. 

While selecting $\beta$\% most uncertain samples at each round, any other data point having the exact same uncertainty as the last element of this set is also added to the uncertainty ground set.

\section{Experiments}
In Table \ref{tab:datasets} we present the details of different datasets used by us in above experiments along with the train-validate split for each.

\begin{table}[ht]
\centering
\begin{adjustbox}{width=1\textwidth}
\small
\begin{tabular}{rlrrrrrrr}
  \hline
 & Name & NumClasses & NumTrain & NumTrainPerClass & NumHoldOut & NumHoldOutPerClass \\ 
  \hline
  &Adience ~\cite{Adience} ~\cite{eidinger2014age} & 2 & 1614 & 790(F), 824(M) & 807 &  395(F), 412(M)   \\
  &MIT-67 ~\cite{MIT67} ~\cite{quattoni2009recognizing} & 67 & 10438 & 68-490 & 5159 &  32-244  \\
  &GenderData & 2 & 2200 & 1087(F), 1113(M) & 548 &  249(F), 299(M)  \\
  &Caltech-101-Mini-1  ~\cite{Caltech-101} ~\cite{fei2007learning} & 10 & 499 & 45-55 & 246 &  20-25  \\
  &Caltech-101-Mini-2  ~\cite{Caltech-101} ~\cite{fei2007learning} & 30 & 1280 & 40-56 & 664 &  15-28 \\
  &CatsVsDogs ~\cite{CatsvsDogs} ~\cite{elson2007asirra} & 2 & 16668 & 8334 & 8332 &  4166 \\
   \hline
\end{tabular}
\end{adjustbox}
\caption{Details of datasets used in our experiments} 
\label{tab:datasets}
\end{table} 

For Goal-1 experiments, we evaluate kNN with k=5. 
For Goal-2 experiments, the number of rounds was experimentally obtained for each experiment as the percentage of data that is required for the model to reach saturation. The values of $B$ and $\beta$ were empirically arrived at and are mentioned in Table \ref{tab:experiments} along with the details about the particular Computer Vision task and dataset used for each set of experiments.

\begin{table}[ht]
\centering
\begin{adjustbox}{width=1\textwidth}
\small
\begin{tabular}{rlrrrrrrr}
  \hline
 & Task & Goal & Dataset & Model & Parameters \\ 
  \hline
  & Gender Recognition & 1 & GenderData & VGGFace/CelebFaces ~\cite{VggFace_CelebFaces} & NA  \\
  & Gender Recognition & 2a & Adience & VGGFace/CelebFaces ~\cite{VggFace_CelebFaces} & $B=0.9,\beta=10$\\
  & Object Recognition & 2a & Caltech-101-Mini-2 & GoogleNet/ImageNet ~\cite{GoogleNet_ImageNet} & $B=1,\beta=10$\\
 & Gender Recognition & 2a & GenderData & VGGFace/CelebFaces ~\cite{VggFace_CelebFaces} & $B=0.12,\beta=10$\\
  & Scene Recognition & 2a & MIT-67 & GoogleNet/Places205 ~\cite{GoogleNet_Places} & $B=2,\beta=10$\\
  & Gender Recognition & 2b & Adience & VGGFace/CelebFaces ~\cite{VggFace_CelebFaces} & $B=5,\beta=10$\\
  & Object Recognition & 2b & Caltech-101-Mini-1 & GoogleNet/ImageNet ~\cite{GoogleNet_ImageNet} & $B=5,\beta=10$\\
  \hline
\end{tabular}
\end{adjustbox}
\caption{Experimental setup} 
\label{tab:experiments}
\end{table} 

In Table\ref{tab:experiments}, the ``Model'' column refers to the pre-trained CNN used to extract features and/or for finetuning. For example, ``GoogleNet/ImageNet'' refers to the GoogleNet model pre-trained on ImageNet data. In the ``Parameters'' column, $B$ and $\beta$ refer to the parameters in FASS and are mentioned in percentages. 

We use Caffe deep learning framework ~\cite{jia2014caffe} for all our experiments.

\section{Results}

\begin{figure*}
\begin{center}
\includegraphics[width = 0.8\textwidth]{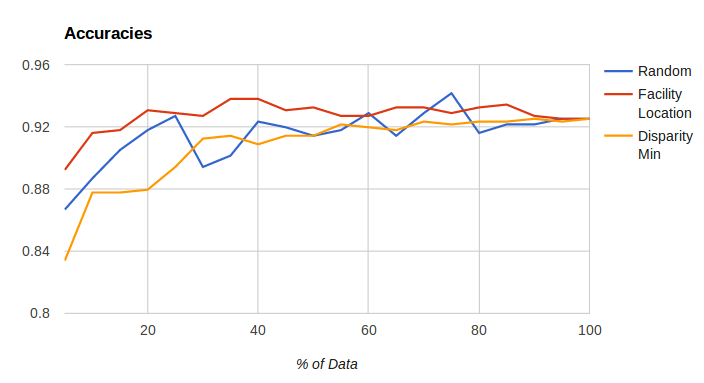}
\end{center}
\caption{Goal-1 (kNN) Results}
\label{fig:goal-1knn}
\end{figure*}

We present Goal-1 experiments results in Figure \ref{fig:goal-1knn}. As hypothesized, using only a portion of training data (about 40\% in our case) we get similar accuracy as with using 100\% of the data. Moreover, Facility Location performs much better than random sampling and Disparity Min, proving that Facility Location function is a good proxy for nearest neighbor classifier. Moreover, this aligns with the theoretical results shown in~\cite{wei2015submodularity}.

\begin{figure*}
\begin{center}
\includegraphics[width = 0.5\textwidth]{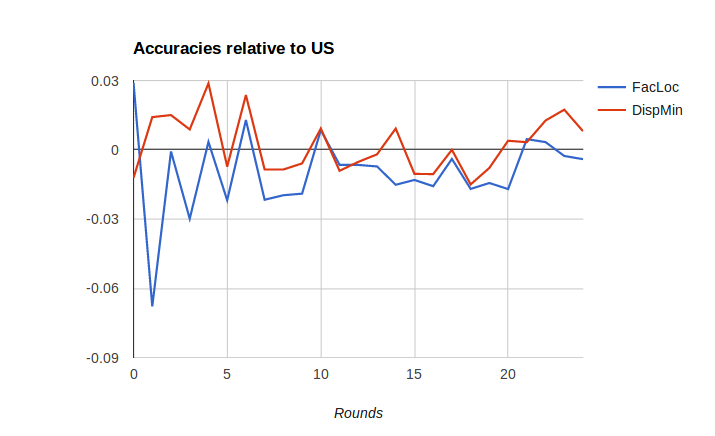}\includegraphics[width = 0.5\textwidth]{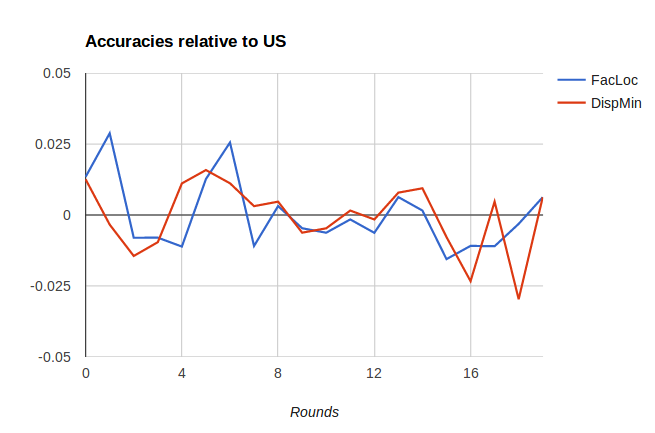}
\end{center}
\caption{Goal-2a (Active Learning - Fine Tuning) Results: Accuracies relative to Uncertainty Sampling (US) Left: Adience Right: Caltech}
\label{fig:goal-2a}
\end{figure*}


For all Goal-2 experiments, we found that the subset selection techniques and uncertainty sampling always outperform random sampling. Moreover, different subset selection techniques work well for different problem contexts, with the intuition behind these results described below.

We present Goal-2a experiments results in Figure \ref{fig:goal-2a}.  Disparity-Min selects most diverse elements which is good for improving model accuracy. Figure \ref{fig:goal-2a} (left) demonstrates the results on Adience while Figure \ref{fig:goal-2a} (right) shows the results on Caltech 101. Both of these are in the context of active learning for fine-tuning. In the case of Finetuning, we observe that our diversity function (disparity min) works better than Facility Location. This can be explained by the fact that Adience has only two classes, and there is a lot of similarity in the face images in this dataset (for example, there is more than one image for the same person), and diversity inherently captures this. In the Caltech dataset however, Facility Location works comparable, or even better than Disparity Min. Since here we have more number of classes, there is already a lot of diversity among the unlabeled points in every round. Thus,it makes sense to select representative images as exemplars for subset selection. Moreover, in both the datasets, we observe that as the number of rounds increase, the diversity/representation among the remaining uncertain unlabeled points reduces and hence neither Facility-Location, nor Disparity-Min fares better than uncertainty sampling.


\begin{figure*}
\begin{center}
\includegraphics[width = 0.5\textwidth]{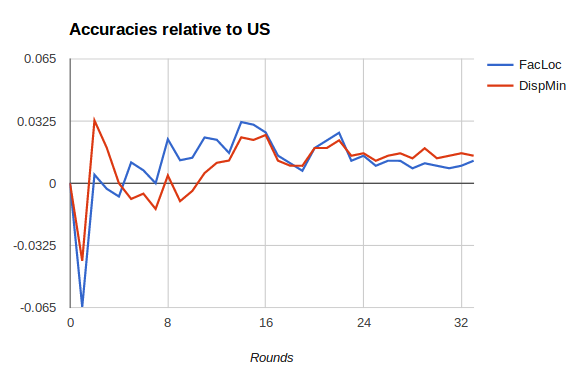}\includegraphics[width = 0.5\textwidth]{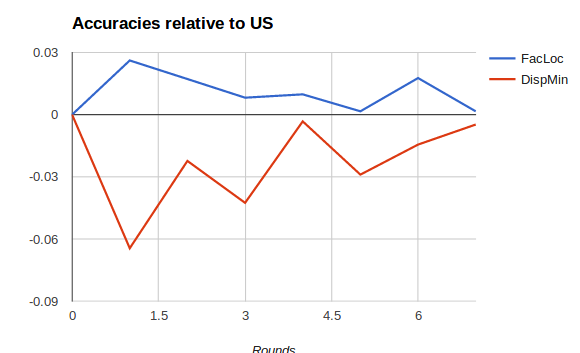}
\end{center}
\begin{center}
\includegraphics[width = 0.5\textwidth]{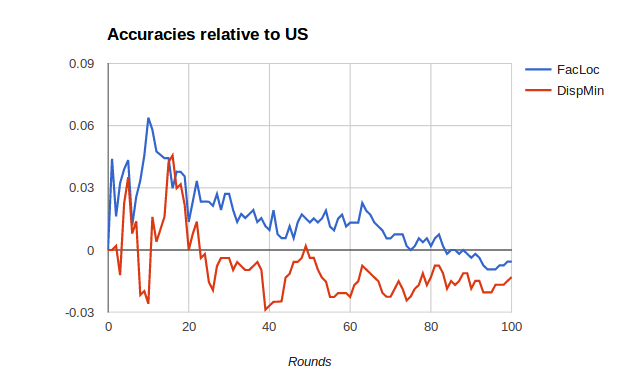}\includegraphics[width = 0.5\textwidth]{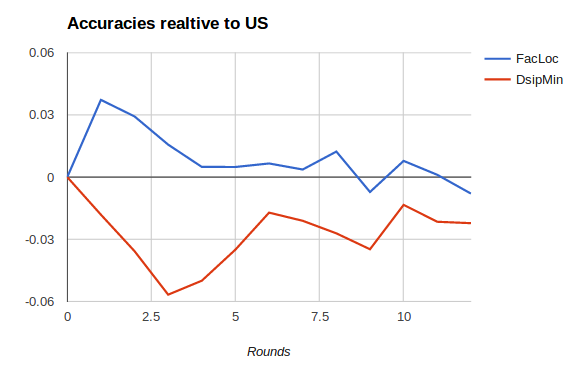}
\end{center}
\caption{Goal-2b Results: Accuracies relative to Uncertainty Sampling (US) Top Left: Adience Top Right: Caltech Bottom Left: Gender Bottom Right: MIT-67}
\label{fig:goal-2b}
\end{figure*}


We present Goal-2b experiments results in Figure \ref{fig:goal-2b}. As pointed out above, Adience data set has lot of diversity within each class and disparity Min performs comparable or even better (initially) compared to Facility Location. On the other hand the other three data sets (Caltech, Gender and MIT67) have a larger number of classes, and the images in each class tend to be more diverse already. Gender refers to a dataset similar to Adience, except that images are more diverse compared (we do not have a repition of faces like the case of Adience). As a result, for the latter three data sets we see Facility-Location performing much better than Disparity-Min. Disparity-Min in this case prefers to choose outliers of each class and hence doesn't help the model. Facility-Location on the other hand, does well to choose representatives from each class, thereby helping the model. In each case, we see that subset selection algorithms on top of Uncertainty Sampling outperform vanilla Uncertainty Sampling and Random Selection. Moreover, the effect of diversity and representation reduce as the number of unlabeled instances reduce in which case Uncertainty sampling already works well. In practice, that would be the point where the accuracy saturates as well.

\section{Conclusion}
Taking the ideas of ~\cite{wei2015submodularity} forward, we demonstrated the utility of subset selection in training models for a variety of standard Computer Vision tasks. Subset selection functions like Facility-Location and Disparity-Min naturally capture the notion of representativeness and diversity which thus help to eliminate redundancies in data. We leverage this and show its practical utility in two settings - ability to learn from less data (Goal-1) and reduced labeling effort (Goal-2). As discussed, these subset selection methods lend themselves well to optimization and hence are easy to implement using simple greedy algorithms. Our results also enabled us to develop insights on the choice of subset selection methods depending on a particular setting. As noted earlier, when ground set is diverse to begin with - either because of diversity \emph{within} a class or diversity due to more number of classes - both Disparity-Min and Facility-Location perform well. On the other hand, when data points within a class are similar, Disparity-Min, in want of picking most mutually diverse images, ends up picking outliers (with respect to items in a class) and confuses the model. We hope that the intuitions obtained from this paper will help practioners develop active learning and subset selection techniques for image recognition tasks. As another potential application, in addition to enabling learning from less data and helping reduce labeling costs, such subset selection techniques can well complement the existing annotation frameworks like ~\cite{vondrick2013efficiently, schoeffmann2015video}.

\bibliographystyle{abbrv}
\bibliography{bmvc}
\end{document}